%% file: main.tex
\documentclass[10pt,twocolumn,letterpaper]{article}

\usepackage{cvpr}              %

\input{preamble}

\definecolor{cvprblue}{rgb}{0.21,0.49,0.74}
\usepackage[pagebackref,breaklinks,colorlinks]{hyperref}

\def\paperID{13882} %
\def\confName{CVPR}
\def\confYear{2024}

\title{Benchmarking Multi-Domain Active Learning \\ on Image Classification}

\author{Jiayi Li\\
Stanford University\\
{\tt\small jiayili@stanford.edu}
\and
Rohan Taori\\
Stanford University\\
{\tt\small rtaori@cs.stanford.edu}
\and
Tatsunori B. Hashimoto\\
Stanford University\\
{\tt\small thashim@stanford.edu}
}

\begin{document}
\maketitle
\input{sec/0_abstract}
\vspace{-1.5em}

\begin{table*}[!htbp]
\centering
\begin{minipage}[c]{0.5\textwidth}
    \begin{subtable}{\textwidth}
        \centering
        \setlength{\tabcolsep}{2pt}
        \begin{tabularx}{\textwidth}{  >{\centering\arraybackslash}X  c  c }
          & Mausoleum & Statue \\
         \rotatebox[origin=c]{90}{Europe} & 
         \begin{minipage}[c]{0.45\textwidth}
            \centering
                \foreach \i in {1,2,3} {
                \includegraphics[width = 0.31\textwidth]{geoyfcc_domain_samples/a\i.jpeg}%
                }
         \end{minipage} & 
         \begin{minipage}[c]{0.45\textwidth}
            \centering
                    \foreach \i in {1,2,3} {
                    \includegraphics[width = 0.31\textwidth]{geoyfcc_domain_samples/c\i.jpeg}%
                    }
         \end{minipage} \\
         [20pt]
         \rotatebox[origin=c]{90}{Asia}  & 
         \begin{minipage}[c]{0.45\textwidth}
            \centering
                \foreach \i in {1,2,3} {
                \includegraphics[width = 0.31\textwidth]{geoyfcc_domain_samples/b\i.jpeg}%
                }
         \end{minipage} & 
         \begin{minipage}[c]{0.45\textwidth}
            \centering
                    \foreach \i in {1,2,3} {
                    \includegraphics[width = 0.31\textwidth]{geoyfcc_domain_samples/d\i.jpeg}%
                }
         \end{minipage}
    \end{tabularx}
    \caption{CLIP-GeoYFCC}
    \end{subtable}
\end{minipage} \hspace{1pt}
\begin{minipage}[c]{0.48\textwidth}
    \begin{subtable}{\textwidth}
    \centering
    \setlength{\tabcolsep}{2pt}
    \begin{tabularx}{\textwidth}{  >{\centering\arraybackslash}X c c }
     & Umbrella & Castle \\
     \rotatebox[origin=c]{90}{Clipart} & 
     \begin{minipage}[c]{0.45\textwidth}
        \centering
            \foreach \i in {1,2,3} {
            \includegraphics[width = 0.31\textwidth]{domainnet_domain_samples/a\i.jpg}%
            }
     \end{minipage} & 
     \begin{minipage}[c]{0.45\textwidth}
        \centering
                \foreach \i in {1,2,3} {
                \includegraphics[width = 0.31\textwidth]{domainnet_domain_samples/c\i.jpg}%
            }
     \end{minipage} \\
     [30pt]
     \rotatebox[origin=c]{90}{Painting}  & 
     \begin{minipage}[c]{0.45\textwidth}
        \centering
            \foreach \i in {1,2,3} {
            \includegraphics[width = 0.31\textwidth]{domainnet_domain_samples/b\i.jpg}%
            }
     \end{minipage} & 
     \begin{minipage}[c]{0.45\textwidth}
        \centering
                \foreach \i in {1,2,3} {
                \includegraphics[width = 0.31\textwidth]{domainnet_domain_samples/d\i.jpg}%
            }
     \end{minipage}
    \end{tabularx}
    \caption{Domainnet}
    \end{subtable}
\end{minipage}
\caption{\textit{Samples of domain differences in CLIP-GeoYFCC and Domainnet}. Rows indicate domains and columns indicate classes. We construct CLIP-GeoYFCC as a new benchmark for multi-domain active learning. Compared with existing domain-focused datasets, CLIP-GeoYFCC is entirely composed of in-the-wild images. In addition, geography-based domains in CLIP-GeoYFCC involve intricate differences in styles and variability in class distributions, which makes it ideal for testing domain-based algorithms.}
\label{fig:sample_pictures}
\vspace{-1em}
\end{table*}

\input{sec/1_intro}

\section{Problem statement}

\subsection{Motivation}
\label{sec:motivation}

While much effort has been invested in improving active learning strategies, few works study how different data characteristics impact the performance of AL methods. 
In fact, by varying the domain composition of the unlabeled data pool,
the same AL method can perform very differently.
In an extreme case in \cref{tab:flip_table}, the same AL method (margin sampling) achieves more than two times the data efficiency of random sampling under one domain composition, but only three-fourths that of random sampling under another composition.  
With a simple change of the domain composition, an AL method could fail and yield worse results than simple random sampling.

This example demonstrates the importance of developing AL methods robust to variations in domain compositions. 
We now discuss the multi-domain active learning problem and evaluation metrics.

\begin{table*}
\begin{tabularx}{\textwidth}{c >{\centering\arraybackslash}X  >{\centering\arraybackslash}X >{\centering\arraybackslash}X }
 \toprule
 Domains & Margin sampling & Random sampling & Data efficiency of margin  \\ 
 \hline
 (a)
\begin{minipage}[c][1cm][c]{0.4\textwidth}
    \includegraphics[width = 0.14 \textwidth]{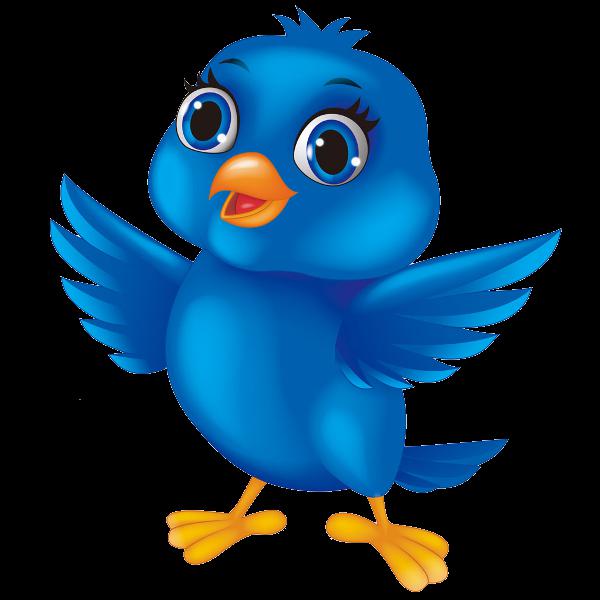}
    \hspace{0.05cm}
    \includegraphics[width = 0.14 \textwidth]{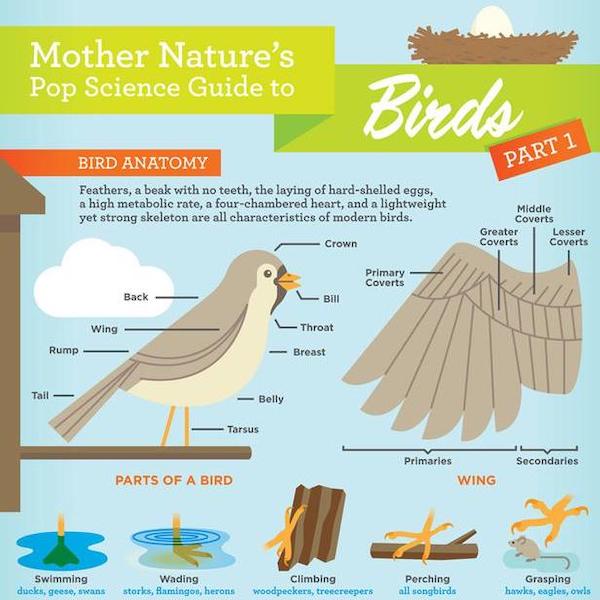}
    \hspace{0.05cm}
    \includegraphics[width = 0.14  \textwidth]{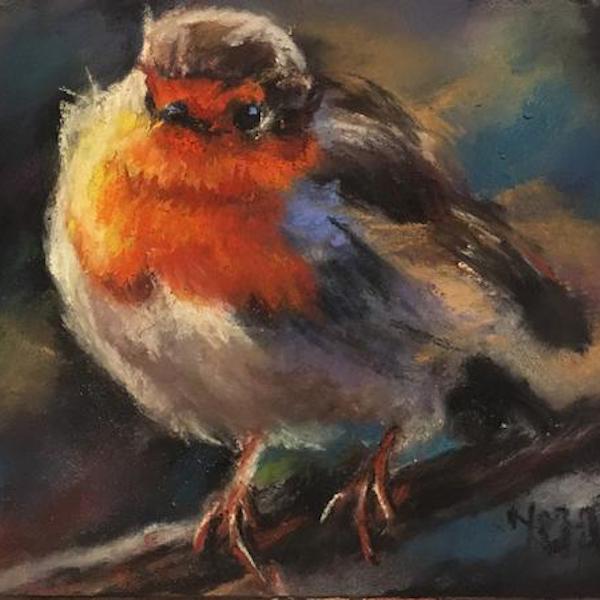}
    \hspace{0.05cm}
    \includegraphics[width = 0.14 \textwidth]{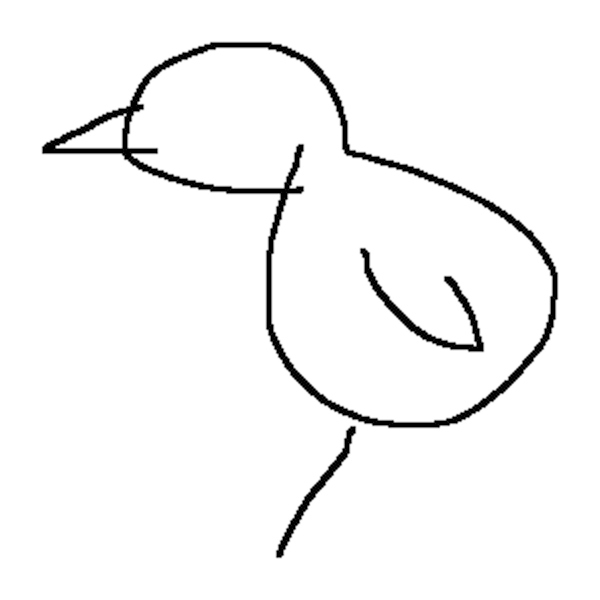}
    \hspace{0.05cm}
    \includegraphics[width = 0.14  \textwidth]{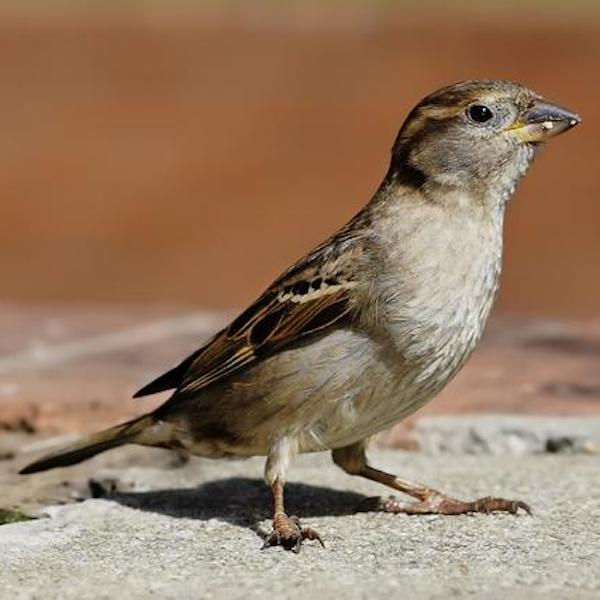}
    \hspace{0.05cm}
    \includegraphics[width = 0.14  \textwidth]{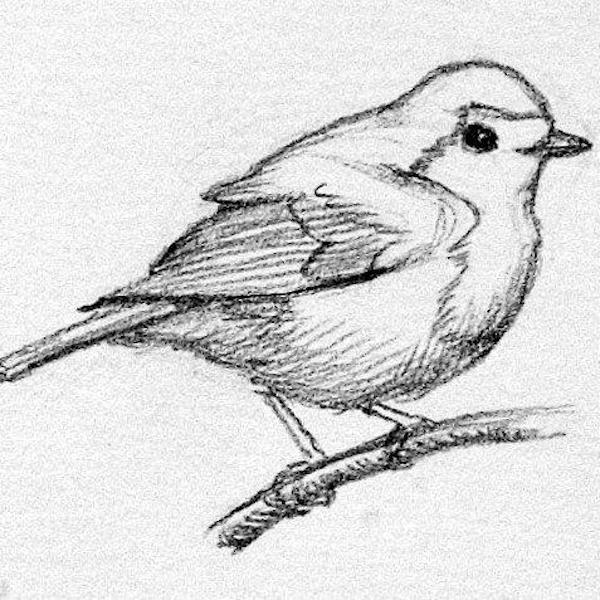}
\end{minipage}
& $\mathbf{69.49}$ & 67.77 & 1.4  \\ 
 \hline
 (b)
 \begin{minipage}[c][1cm][c]{0.4\textwidth}
    \includegraphics[width = 0.14 \textwidth]{domainnet_domain_samples/clipart.jpg}
    \hspace{0.05cm}
    \includegraphics[width = 0.14  \textwidth]{domainnet_domain_samples/painting.jpg}
    \hspace{0.05cm}
    \includegraphics[width = 0.14 \textwidth]{domainnet_domain_samples/quickdraw.jpg}
    \hspace{0.05cm}
    \includegraphics[width = 0.14  \textwidth]{domainnet_domain_samples/real.jpg}
    \hspace{0.05cm}
    \includegraphics[width = 0.14  \textwidth]{domainnet_domain_samples/sketch.jpg}
\end{minipage}
 & $\mathbf{73.72}$ & 71.04 & 2.1 \\
 \hline
 (c)
 \begin{minipage}[c][1cm][c]{0.4\textwidth}
    \includegraphics[width = 0.14 \textwidth]{domainnet_domain_samples/clipart.jpg}
    \hspace{0.05cm}
    \includegraphics[width = 0.14 \textwidth]{domainnet_domain_samples/infograph.jpg}
    \hspace{0.05cm}
    \includegraphics[width = 0.14 \textwidth]{domainnet_domain_samples/quickdraw.jpg}
    \hspace{0.05cm}
    \includegraphics[width = 0.14  \textwidth]{domainnet_domain_samples/real.jpg}
\end{minipage}
  & 69.18 & $\mathbf{69.79}$ & \textcolor{red}{\textbf{0.75}}  \\
 \bottomrule
\end{tabularx}
\smallskip
\caption{\textit{Test accuracies (\%) of margin and random sampling on different Domainnet domain compositions}. Experiment (a) contains all six domains in Domainnet which are clipart, infograph, painting, quickdraw, real and sketch in order of the figures. Experiments (b) and (c) contain 5 and 4 domains respectively. Margin sampling achieves higher test accuracies than random sampling in (a) and (b) but performs worse than random sampling in (c). Definition of data efficiency is stated in \cref{de}; experiment details are described in \cref{sec:domain_comp_setup}. Accuracies are measured at query round 20.}
\label{tab:flip_table}
\vspace{-1em}
\end{table*}

\subsection{Setup}
\label{sec:setup}

We assume there exists a large pool of unlabeled data $U = \{x_i\}$ with known domains. A seed set of $m_0$ samples are uniformly drawn from $U$ and are labeled to form the labeled training set $L = \{(x_i, y_i)\}$. We assume that the unlabeled pool $U$ consists of samples from $N$ disjoint domains, denoted by $U_1, ...,  U_N$. For each domain $j$, we have a validation set $V_j$. 

A model $M$ is trained on the initial $L$ and is evaluated on $V_1, ..., V_N$.
In each round $r$, the AL algorithm leverages $M$ and validation results to select $m$ samples $\{x_i\}$ from $U \setminus L$ to label and the labeled batch $B = \{(x_i, y_i)\}$ is added to the existing $L$. We use $m_i^r$ to denote the labeling budget for domain $i$ in round $r$, and so $m = \sum_i m_i^r$. Model $M$ is retrained from scratch on $L$ for the next query round. 

We assume abundant computation resource and focus on investigating sampling strategies that achieve highest validation accuracies with the fewest labeled samples.

\subsection{Evaluation metrics}
\label{sec:metrics}

For each domain $i$, we label a small number of samples to serve as the validation set $V_i$. 
The test accuracy of model $f$ on domain $i$ is evaluated on the validation set $V_i$ and is denoted as $a_i(f)$. 

When evaluating the efficacy of active learning strategies, it is most natural to use a validation set $V$ with the same data distribution as the unlabeled training set $U$. The in-distribution validation accuracy, or \emph{ambient accuracy}, is defined as 
\begin{equation}
    a_{ambient} (f) = \sum_i \frac{|U_i|}{|U|} a_i(f).
\end{equation}
The \emph{mean-group accuracy} across domains gives each domain the same weight and evaluates the average performance of the trained model: 
\begin{equation}
    a_{mean} (f) = \sum_i \frac{1}{N} a_i(f).
\end{equation}
The \emph{worst-group accuracy} measures how robust the model is in the most difficult domain:
\begin{equation}
    a_{worst} (f) = \min_i a_i(f).
\end{equation}

Among the three metrics, ambient accuracy forecasts the performance of the model on test data whose domain distribution is the same as the training data. Mean-group accuracy and worst-group accuracy estimate how robust the trained model is under domain distribution shift. All three metrics need to be considered in multi-domain active learning because an AL method that performs well on one metric often faces severe compromise on other metrics. We further discuss this trade-off between metrics in \cref{sec:tradeoff}. 

\begin{table*}[t]
  \centering
  \begin{tabularx}{\textwidth}{l|rrrY}
    \toprule
    Dataset & Images & Classes & Domains & Domain Classification Criteria \\
    \hline
    Office-Caltech \cite{Office-Caltech} & 2,533 & 10 & 4 & Sources (web-cam, Amazon etc.) \\
    Office-Home \cite{Office-Home} & 15,588 & 65 & 4 & Genres of images \\
    Digits-Five & ~100,000 & 10 & 5 & Styles of digits \\
    Domainnet-345 \cite{domainnet} & 586,575 & 345 & 6 & Genres of images \\
    \textbf{CLIP-GeoYFCC-350} & \textbf{338,730} & \textbf{350} & \textbf{5} & \textbf{Geographical origins of images}\\
    \bottomrule
  \end{tabularx}
  \smallskip
  \caption{\textit{Comparing commonly used domain-based image datasets to our proposed dataset (last line in \textbf{bold})}. Digits-Five dataset contains five digit datasets (MNIST \cite{MNIST}, MNIST-M \cite{MNIST-M}, Synthetic Digits \cite{MNIST-M}, SVHN \cite{SVHN}, and USPS). CLIP-GeoYFCC contains a large number of images and classes. More importantly, domain classification based on geographical origins presents new challenges as it involves complicated differences in styles, cultures, and class distributions.}
  \label{tab:dataset_info}
\end{table*}

\section{Domain-based benchmark for active learning}
\label{sec:benchmark}
We develop a new version of GeoYFCC \cite{geoyfcc} to facilitate future research on multi-domain AL. 
The original GeoYFCC collects images with geotags (latitude and longitude) from YFCC100M \cite{yfcc}.
Images are assigned labels based on filtering keywords of image tags;
however, we discover that GeoYFCC contains much label noise as images with the same label show very different subjects. Since GeoYFCC uses keyword filtering to acquire labels, it is likely that the hard-matched labels do not directly describe the subjects of the images. 

In order to obtain more accurate labels for the images, we use CLIP \cite{CLIP} to relabel the GeoYFCC dataset. We calculate image features and match them with labels in ImageNet-21K that have the closest text features. Geotags are used to assign each image to its corresponding continent and this results in six domains that are Africa, Asia, Europe, North America, South America, and Oceania. We select continents instead of countries to be the domains because countries geographically close to each other demonstrate insignificant domain difference. 
This results in a dataset with 1,146,768 images from 18814 classes and 6 continents. We refer to this dataset as CLIP-GeoYFCC . 

CLIP-GeoYFCC is a large real-world dataset with images from diverse categories. To the best of our knowledge, the only other domain-based dataset that matches its scale and diversity is Domainnet \cite{domainnet}, which contains 586,575 images from 345 classes and 6 domains.
We further filter and balance CLIP-GeoYFCC (\cref{app:filter_geoyfcc}),
resulting in a 350 class subset with 338,730 images, which we call CLIP-GeoYFCC-350 (\cref{tab:dataset_info}).
From this, we further subsample 45 classes to form CLIP-GeoYFCC-45 containing 86,288 images.
These two datasets mirror two publicly released Domainnet versions, Domainnet-345 with 345 classes and Domainnet-40 with 40 classes and 72,614 samples. 

\def\cca#1{%
    \pgfmathsetmacro\calc{(#1-0)*100/(80-0)}%
    \edef\clrmacro{\noexpand\cellcolor{blue!\calc}}%
    \clrmacro%
    \ifdim \calc pt>50pt\color{white}\fi{#1}%
}
\begin{table*}[t!]
\centering

\begin{subtable}[t]{0.48\textwidth}
    \centering
    \caption{CLIP-GeoYFCC-350}
    \begin{tabularx}{\textwidth}{l|YYYYY}
            \toprule
                      & Africa        & Asia          & EU.        & N.A. & S.A. \\ \hline
            Africa        & \textbf{\cca{56.0}} & \cca{12.3}          & \cca{11.2}          & \cca{16.5}          & \cca{9.9}           \\
            Asia          & \cca{18.0}          & \textbf{\cca{37.0}} & \cca{17.9}          & \cca{22.3}          & \cca{14.9}          \\
            EU.        & \cca{10.5}          & \cca{15.6}          & \textbf{\cca{38.8}} & \cca{26.5}          & \cca{18.5}          \\
            N.A. & \cca{13.6}          & \cca{15.6}          & \cca{23.3}          & \textbf{\cca{48.0}} & \cca{21.9}          \\
            S.A. & \cca{11.8}          & \cca{14.1}          & \cca{20.4}          & \cca{27.7}          & \textbf{\cca{45.7}} \\ \bottomrule
        \end{tabularx}
\end{subtable}%
\hfill %
\begin{subtable}[t]{0.5\textwidth}
    \centering
    \caption{Domainnet-345}
    \begin{tabularx}{\textwidth}{l|YYYYYY}
 \toprule
                      & Clip    & Info    & Paint      & Quick   & Real        & Sketch          \\ \hline
            Clip     & \textbf{\cca{60.8}} & \cca{8.8}          & \cca{21.4}          & \cca{6.1}          & \cca{35.2}        & \cca{26.3}          \\
            Info   & \cca{13.7}          & \textbf{\cca{21.5}} & \cca{14.2}          & \cca{1.2}          & \cca{21.6}        & \cca{11.3}          \\
            Paint    & \cca{22.2}          & \cca{9.0}          & \textbf{\cca{53.1}} & \cca{1.1}          & \cca{39.2}        & \cca{20.1}          \\
            Quick    & \cca{5.7}           & \cca{1.0}        & \cca{1.0}          & \textbf{\cca{53.1}} & \cca{2.1}    & \cca{5.3}          \\
            Real   & \cca{26.6}          & \cca{10.1}          & \cca{30.6}         & \cca{2.3}          & \textbf{\cca{66.7}} & \cca{17.8} \\ 
            Sketch  & \cca{27.4}          & \cca{6.6}          & \cca{17.8}         & \cca{7.0}          & \cca{26.3} & \textbf{\cca{50.1}}\\\bottomrule
    \end{tabularx}
\end{subtable}

\smallskip
\caption{\textit{Cross domain accuracy (\%) for CLIP-GeoYFCC-350 and Domainnet-345}. In (a), EU., N.A. and S.A. stand for Europe, North America and South America, respectively. In each table, the rows represent the source domains and the columns represent the target/validation domains. Training set size is 10k for both tables. }
\label{tab:geoyfcc_transfer}
\vspace{-1em}
\end{table*}

Stylistic differences between Domainnet and CLIP-GeoYFCC are also reflected in experimental statistics. To understand relationships and similarities between domains, we run experiments on both Domainnet and CLIP-GeoYFCC where a model is only trained on data from one domain and tested on the other domains. As shown in \cref{tab:geoyfcc_transfer}, there exist domains in Domainnet that transfer little useful information to other domains. For example, training on quickdraw achieves only 1\% validation accuracy on inforgraph. In contrast, CLIP-GeoYFCC provides naturalistic domains based on geographical locations with a more pervasive transfer across domains. 
These two datasets provide complementary challenges for any AL algorithm and thus form the basis for our benchmark.

\section{Active learning methods}

In multi-domain active learning, apart from applying the standard active learning methods on the entire dataset, we investigate  hierarchical approaches where domain structures are explicitly considered. For each query round, we first allocate labeling budget for each domain and then within each domain, we apply the instance-level query strategies to select the dictated number of samples (See \cref{sec:alg} for details).

\subsection{Instance-level query strategy}
\label{sec:standard_methods}

We use margin sampling \cite{margin1} 
\footnote{We leave benchmarking clustering-based AL strategies like BADGE \cite{BADGE} and ClusterMargin \cite{batchAL} for future work. 
As discussed in \cref{sec:related_work}, most existing benchmarking work on AL has found that classical margin sampling is a strong, if not often dominant, baseline \cite{bahri2022margin}. 
Thus, we treat margin sampling as our reference point to explore and benchmark domain-aware AL strategies} 
as single-domain AL baseline and as the instance-level query strategy for various domain allocations. Margin sampling chooses samples with the smallest differences in model-predicted probabilities between the first and second most likely classes. Mathematical definition of margin and comparison with other uncertainty-based methods are included in \cref{sec:margin_best}. 

\subsection{Domain allocation strategies}

We benchmark four domain allocation strategies: uniform allocation, error-proportional allocation, loss-exponential allocation, and worst-group allocation (see \cref{sec:domain_allocation} for mathematical definitions).

These four domain allocation strategies are chosen to form a spectrum with respect to prioritizing samples from the worst group. 
In \cref{sec:tradeoff}, we evaluate these strategies across different metrics and find very different performance characteristics and tradeoffs across the metrics for each. 

\section{Experiments: effect of domain composition on learning efficiency}
\label{sec:domain_effect}

\begin{figure*}
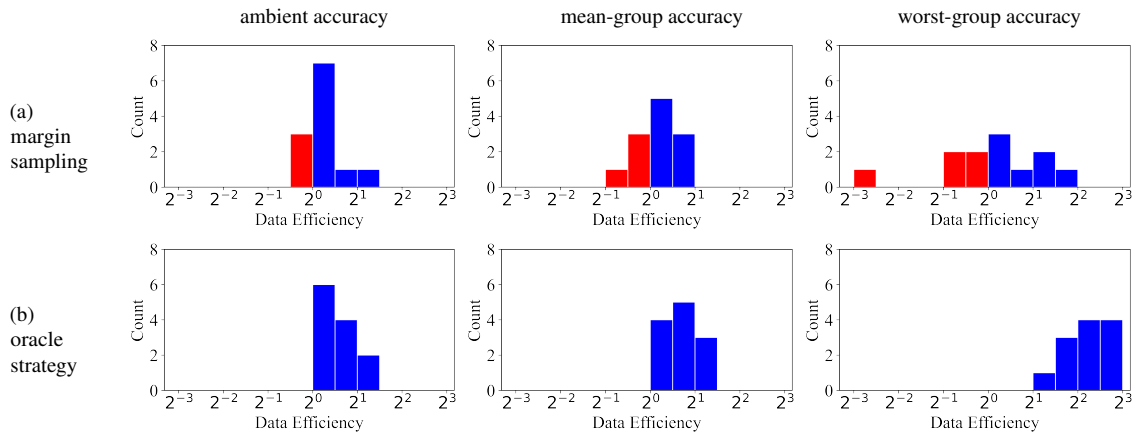

\centering
\begin{minipage}[c]{0.07\textwidth}
\end{minipage}\hfill%
\begin{minipage}[c]{0.9\textwidth}
\foreach \accuracy in {ambient,mean-group,worst-group} {\centering
        \begin{subfigure}{0.33\linewidth}
            \caption*{\qquad\accuracy\ accuracy}
        \end{subfigure}%
    }%
\end{minipage}

\begin{subfigure}{\linewidth}\centering%
\begin{minipage}[c]{0.07\textwidth}
\begin{subfigure}{\linewidth}%
\caption{\\margin sampling}%
\end{subfigure}%
\end{minipage}\hfill%
\begin{minipage}[c]{0.9\textwidth}
    \foreach \accuracy in {ambient,mean-group,worst-group} {\centering%
        \begin{subfigure}{0.33\linewidth}%
            \includegraphics[width=1.0\textwidth]{figures/domain_effect_margin_sampling_\accuracy.png}%
        \end{subfigure}%
    }%
\end{minipage}%
\end{subfigure}

\begin{subfigure}{\linewidth}\centering%
\begin{minipage}[c]{0.07\textwidth}
\begin{subfigure}{\linewidth}%
\caption{\\oracle strategy}%
\end{subfigure}%
\end{minipage}\hfill%
\begin{minipage}[c]{0.9\textwidth}
    \foreach \accuracy in {ambient,mean-group,worst-group} {\centering%
        \begin{subfigure}{0.33\linewidth}%
            \includegraphics[width=1.0\textwidth]{figures/domain_effect_oracle_strategy_\accuracy.png}%
        \end{subfigure}%
    }%
\end{minipage}%
\end{subfigure}
    \caption{Comparison of margin sampling (top row) and oracle strategy (bottom row) on Domainnet-40 domain compositions. Data efficiencies are calculated after 20 rounds of query.
    The $x$ axis is drawn in log scale.
    }
    \label{fig:domain_effect}
    \vspace{-1em}
\end{figure*}

Through experiment (c) in \cref{tab:flip_table}, we presented a domain composition where margin sampling achieves lower ambient accuracy than random sampling. 
To examine the failure of margin sampling in more generality (across different domain compositions and different evaluation metrics), we expand our investigation;
we sample 12 domain compositions and evaluate experiments using all three metrics. 
While single-domain margin sampling sometimes fails to provide gain over random sampling, we wonder how much improvement is achievable utilizing domain structures.  
We investigate the optimal performance achievable by domain allocation strategies by constructing a heuristic oracle strategy, which for each domain composition takes the winner from 6 different domain allocation strategies. 

\paragraph{Experiment setup}
\phantomsection
\label{sec:domain_comp_setup}
From the six domains contained in Domainnet-40, we randomly sample 3 to 6 domains to form the data pools. For example, in experiment (c) of \cref{tab:flip_table}, the unlabeled data pool contains four domains, namely clipart, infograph, quickdraw, and real. We sampled 12 domain combinations in total and each histogram in \cref{fig:domain_effect} presents results of 12 experiments. 

In \cref{fig:domain_effect}, oracle strategies refer to strategies constructed for each domain combination by taking the best performing algorithm among random sampling, margin sampling, uniform margin, error-proportional margin, loss-exponential margin, and worst-group margin. All oracle strategies use margin sampling as the instance-level query method.

Other experiment details and hyperparameters are listed in \cref{sec:experiment_details}.

\paragraph{Result}

We observe that in the multi-domain scenarios, applying margin sampling without considering the domain structures could lead to data efficiencies less than 1 (marked by red in \cref{fig:domain_effect}). In specific, failure of margin sampling occurs in 3/12, 4/12, and 5/12 experiments when evaluated under ambient, mean-group, and worst-group accuracies respectively. The fact that single-domain AL solution could often lead to unwanted results in multi-domain datasets indicates the necessity of more investigation into multi-domain active learning.

By utilizing the domain structures, oracle strategies achieve data efficiencies larger than 1 under all domain compositions. In addition, oracle strategies improve significantly upon the performance of margin sampling by an average of 0.08, 0.39, and 3.49 in data efficiency in terms of ambient, mean-group, and worst-group accuracies respectively. 

We note that the oracle strategy relies on extra information unavailable to practical agents. However, this result still demonstrates that there is huge room for improving performance using domain structures and calls for more research about domain allocation strategies, which we explore next.

\section{Experiments: benchmarking domain allocation strategies}

The experiments in \cref{sec:domain_effect} demonstrate the necessity of investigating multi-domain active learning and the potential benefit of using explicit domain-level allocations. In this section, we compare various domain-level allocation strategies on CLIP-GeoYFCC and Domainnet, our two benchmark multi-domain datasets (\cref{sec:benchmark}). 
We are interested in exploring three specific research questions:
\begin{itemize}
    \item Firstly, given the failures of single-domain margin sampling illustrated in \cref{sec:domain_effect}, are there simple domain allocation strategies that consistently improve upon margin sampling on specific metrics? 
    \item Secondly, given a metric, could we find a strategy that outperforms other strategies on all datasets? 
    \item Lastly, do domain allocation strategy excelling in one metric also achieve good performance on other metrics? 
\end{itemize}
Our benchmarking effort provides some preliminary findings on these questions and signals the need for more research.
Experiment details and hyperparameters are listed in \cref{sec:experiment_details}.

\subsection{Room for improvement on individual metrics}
In our benchmarking, we found that simple domain allocations strategies improve upon margin sampling on mean-group and worst-group metrics. When evaluated under mean-group accuracy, uniform and error-proportional domain allocations achieve higher test accuracies than margin sampling on all datasets (\cref{fig:four_datasets_onlymargin}). In specific, uniform\_margin achieves 0.96\% higher mean-group accuracy than margin sampling on average. When we look at worst-group accuracy, worst-group allocation leads margin sampling by an average of 3.6\% (\cref{aggregate_table}).

On ambient accuracy, there is no simple domain allocation strategy that provides consistent improvement upon margin sampling. In \cref{sec:threshold}, we describe one variation of the margin method we discovered called threshold-margin.
Threshold-margin achieves slightly better ambient performance than margin sampling in aggregate; algorithm details and initial results are provided in the appendix. 

\subsection{Optimal strategy on a given metric}
\label{sec:optimal}

\begin{figure*}[h]
    \centering
    \includegraphics[width=\textwidth]{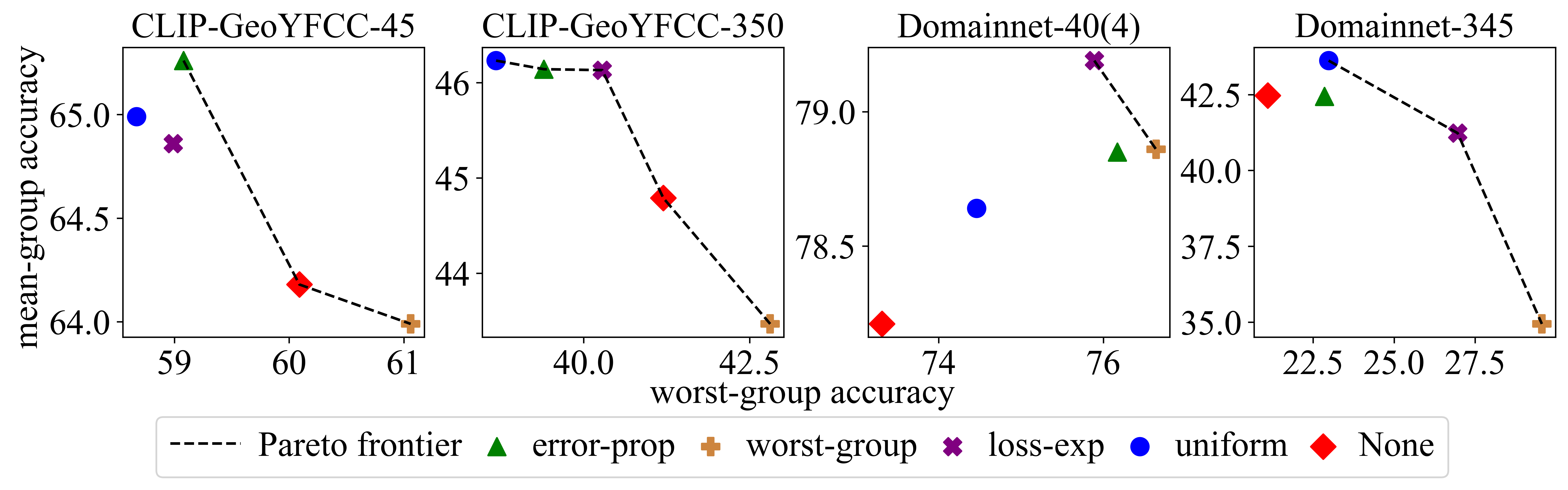}
    \caption{\textit{Comparing different domain allocation methods on CLIP-GeoYFCC-45, CLIP-GeoYFCC-350, Domainnet-40(4), and Domainnet-345}. Domainnet-40(4) consists of 4 domains (clipart, painting, real, sketch) out of the 6 domains in Domainnet-40. 
    All methods use margin sampling as the instance-level query strategy.
    Accuracies are evaluated after 40 rounds of query. 
    The dashed line represents the Pareto frontier \cite{Martinez2020pareto} between the mean-group and worst-group accuracies.
    }
    \label{fig:four_datasets_onlymargin}
\vspace{-1em}
\end{figure*}

While simple domain allocations can improve upon margin sampling on some metrics, we observe that no single domain allocation strategy always dominates mean-group and ambient accuracies. For example, although uniform domain allocation achieves higher mean-group accuracies than margin sampling on all datasets, in mean-group accuracies error-proportional and loss-exponential domain allocations are sometimes the winner (\cref{fig:four_datasets_onlymargin}). In other words, the choice of optimal domain allocation strategies depends on domain dynamics of different datasets. 

The only outlier is the worst-group metric, for which worst-group allocation has an obvious advantage across all datasets.

\begin{table}[]
\centering
\begin{tabularx}{1.0\columnwidth}{l YYY}
\toprule
\textbf{method} &
  \textbf{ambient} &
  \textbf{mean group} &
  \textbf{worst group} \\
  \midrule
margin              & \textbf{58.66}         & 57.41          & 48.92         \\
uniform margin     & \underline{58.49}         & \textbf{58.37} & 48.70          \\
error-prop margin & 58.24          & \underline{58.17}          & 49.37          \\
loss-exp margin   & 57.85          & 57.85         & \underline{50.53}          \\
worst-group margin      & 55.97          & 55.32          & \textbf{52.52} \\
\bottomrule
\end{tabularx}
\caption{\textit{Aggregate accuracy (\%) of different domain allocation methods}. Reported accuracies are averaged over the experiments on CLIP-GeoYFCC-45, CLIP-GeoYFCC-350, Domainnet-40(4), and Domainnet-345.
The \textbf{highest} and \underline{second highest} accuracies are shown in \textbf{bold} and \underline{underline}, respectively.
Readers are referred to \cref{fig:four_datasets_onlymargin} for results on individual datasets.
}
\label{aggregate_table}
\end{table} 

\subsection{Tradeoff between metrics}
\label{sec:tradeoff}

We observe that in multi-domain active learning, methods excelling in one metric often suffer from severe compromise on other metrics. For example, worst-group allocation dominates on worst-group accuracy, but it often achieves significantly lower mean-group and ambient accuracies than other domain allocation strategies. On average, worst-group allocation is 1.88\% and 2.53\% lower than the second worst domain allocation strategy on ambient and mean-group accuracies respectively (\cref{aggregate_table}).

In addition, different datasets have very different charateristics of tradeoff between metrics. For Domainnet, four out of the six domains (clipart, painting, real, and sketch) are very similar to each other while the other two domains (quickdraw and infograph) are drastically different from the rest (\cref{tab:geoyfcc_transfer}). Because of the unusually low transfer between some domains, the worst-group allocation that samples mainly from infograph on Domainnet-345 suffers from huge compromise on mean-group accuracy (9\% lower than uniform allocation). When we remove quickdraw and infograph, the correlation between worst-group and mean-group metrics is completely reversed (see Domainnet-40(4) in \cref{fig:four_datasets_onlymargin}). 

As a comparison, CLIP-GeoYFCC does not involve domains as unusual as infograph, and the geography-based domains demonstrate moderate and pervasive transfer between each other (\cref{tab:geoyfcc_transfer}). As a result, on CLIP-GeoYFCC-350, worst-group allocation is only 2.5\% lower then uniform allocation on mean-group accuracy. In addition, removing some of the domains in CLIP-GeoYFCC would not dramtically change the tradeoff behavior. Thus, CLIP-GeoYFCC provides more natural, complementary environment to evaluate domain allocation strategies' robustness to domain differences.

\section{Related work}
\label{sec:related_work}

\paragraph{Uncertainty-based Sampling.}
Uncertainty-based sampling is the most successful and widely-used query framework in AL  \cite{Mussmann2018, Mussmann2020, peng2021, pretrain, BASE}. Uncertainty-based methods query instances that the current model is least certain about \cite{Settles2010, lewis1995}. Popular uncertainty measures include confidence \cite{LC}, margin \cite{margin1, margin2}, and entropy \cite{entropy}.

On top of the uncertainty-based framework, researchers have investigated using clustering to increase the diversity of selected instances, in order to learn an embedding representative of the entire dataset \cite{BADGE, CORESET, batchAL}. 
While many such AL strategies have been proposed, \cite{bahri2022margin, gissin2019discriminative, ALbert} showed that classic margin sampling matches or outperforms all others, including current state-of-the-art, in a wide range of experimental settings. 
\cite{bahri2022margin} explicitly calls for AL benchmarking against margin sampling. 

\paragraph{Active Domain Adaptation.}
While traditional AL deals with single domain data \cite{BASE, BADGE,CORESET, bahri2022margin}, active domain adaptation (ADA) tackles problems with multiple domains. ADA is an extension of domain adaptation (DA) which aims to transfer learning on one source domain to a different target domain \cite{Prabhu2021activeadapt,su2020domainadapt}. In ADA, AL is used to select useful samples from the unlabeled target domain. Our work considers multi-domain active learning (MDAL) which has only been studied recently \cite{hierarchical, he2022multidomain}. The key difference between MDAL and ADA is that for MDAL, both training data and target test data contain multiple domains, which poses bigger challenge than domain-to-domain adaptation.   

\paragraph{Multi-domain active learning.}
\cite{he2022multidomain} approached multi-domain active learning from a domain adaptation perspective where domain information is used during model training but not in instance sampling. We point out that on all experiments using deep neural networks in \cite{he2022multidomain}, domain adaptation schemes \cite{MAN, DANN, CAN, MDNet} do not  improve upon the baseline of training a single model on the mixed data.
Our approach to MDAL is very similar to \cite{hierarchical} in that domain information is used in a two-step instance sampling process. 

Our work differs from existing MDAL investigations in two aspects. Firstly, no previous work comprehensively evaluates domain allocation strategies on ambient, mean-group, and worst-group metrics. \cite{Sharaf2022parity,  Abernethy2022minmax, Shekhar2021, Abernethy2020disparate, Anahideh2022fairAL} focus on worst-group accuracy and investigates variations of worst-group allocation. \cite{hierarchical, he2022multidomain} evaluate average accuracies over domains.  
Secondly, previous works only investigated image datasets with genre-based domains which are limited in number of classes and variability of domains. We propose CLIP-GeoYFCC as a non-genre-based multi-domain dataset for more robust evaluation.

\section{Conclusion}

In this work, we demonstrate that the multi-domain nature of real world data drastically challenges the single-domain assumptions of existing AL algorithms.
To advance research on AL algorithms robust to diverse domain compositions, we introduce a novel benchmark dataset, CLIP-GeoYFCC. CLIP-GeoYFCC supplements existing AL datasets by providing naturalistic geography-based domains.
Through benchmarking on CLIP-GeoYFCC and Domainnet, we draw attention to two important open question: can we develop domain allocation strategies that work well across \emph{different domain compositions}, and across \emph{different metrics}?

A limitation of this work is that we only evaluate data selection methods. Evaluation of different learning algorithms is left to future work.

{
    \small
    \bibliographystyle{ieeenat_fullname}
    \bibliography{main}
}

\input{sec/X_suppl}

\end{document}

%% file: preamble.tex
\usepackage[dvipsnames]{xcolor}
\usepackage[utf8]{inputenc} %
\usepackage[T1]{fontenc}    %
\usepackage{url}            %
\usepackage{booktabs}       %
\usepackage{amsfonts}       %
\usepackage{nicefrac}       %
\usepackage{microtype}      %

\usepackage{geometry}
\usepackage{graphicx}
\usepackage{subcaption}
\usepackage{caption}
\usepackage{float}
\usepackage{pgffor}
\usepackage{pifont}
\usepackage{sidecap}
\usepackage{tikz}
\sidecaptionvpos{figure}{c}
\usepackage{tabularx} 
\newcolumntype{Y}{>{\centering\arraybackslash}X}
\usepackage{subcaption}
\usepackage{ragged2e}
\usepackage{array}  
\usepackage{array,multirow,graphicx}

\usepackage{xargs}          %

\usepackage[disable]{todonotes} %
\colorlet{rt_color}{gray}
\colorlet{jl_color}{purple}
\definecolor{th_color}{rgb}{0.0, 0.5, 0.0}
\newcommandx{\rtnote}[2][1=]{\todo[linecolor=rt_color,backgroundcolor=rt_color!25,bordercolor=rt_color,#1]{R: #2}}
\newcommandx{\jlnote}[2][1=]{\todo[linecolor=jl_color,backgroundcolor=jl_color!25,bordercolor=jl_color,#1]{J: #2}}
\newcommandx{\thnote}[2][1=]{\todo[linecolor=th_color,backgroundcolor=th_color!25,bordercolor=th_color,#1]{T: #2}}
\newcommandx{\thc}[2][1=]{\todo[linecolor=th_color,backgroundcolor=th_color!25,bordercolor=th_color,#1]{T: #2}}

\usepackage{amsthm}
\usepackage{xcolor}
\usepackage{color}
\usepackage{amsmath}

\usepackage{colortbl}
\usepackage{calc}

\DeclareMathOperator*{\argmax}{arg\,max}
\DeclareMathOperator*{\argmin}{arg\,min}

\usepackage{algorithm}
\usepackage{algpseudocode}
\usepackage{bbm}

\newlength{\tempht}
\newsavebox{\tempbox}
\newcommand{\mysameheightbox}[2]{{%
    \setlength{\tempht}{5cm} %
    \sbox\tempbox{%
      \resizebox{#1}{!}{%
        #2 %
      }} %
    \setlength{\tempht}{\ht\tempbox} %
    #2 %
  }}

%% file: sec/0_abstract.tex
\begin{abstract}

Active learning aims to enhance model performance by strategically labeling informative data points.
While extensively studied, its effectiveness on large-scale, real-world datasets remains underexplored.
Existing research primarily focuses on single-source data, ignoring the multi-domain nature of real-world data.
We introduce a multi-domain active learning benchmark to bridge this gap. 
Our benchmark demonstrates that traditional single-domain active learning strategies are often less effective than random selection in multi-domain scenarios. 
We also introduce CLIP-GeoYFCC, a novel large-scale image dataset built around geographical domains, in contrast to existing genre-based domain datasets.
Analysis on our benchmark shows that all multi-domain strategies exhibit significant tradeoffs, with no strategy outperforming across all datasets or all metrics,
emphasizing the need for future research.
\end{abstract}

%% file: sec/1_intro.tex
\section{Introduction}
\label{sec:intro}

Many successes in machine learning over the past decade
have been due to careful data annotation efforts,
in both training from scratch \cite{imagenet} and fine-tuning pre-trained models \cite{huh2016}. 
However, in practice, data labels are often expensive and time-consuming to obtain.
Active learning (AL) promises to significantly reduce annotation
costs by adaptively selecting the most informative datapoints from a pool of candidates \cite{Settles2010}.
However, despite huge research efforts \cite{Ren2022}, 
evaluation of AL methods on large-scale, in-the-wild data is lacking. 
Existing AL methods focus on single domain data, with very few investigations in multi-domain scenarios \cite{hierarchical}.

We first evaluate standard single-domain AL strategies on multi-domain datasets.
We discover that the performance is highly dependent on domain composition.
The data efficiency (defined in \cref{de}) of the same algorithm can vary from 0.75x (\emph{worse than random sampling}) to 2.1x for different domain compositions (\cref{tab:flip_table}). These findings demonstrate that the multi-domain nature of real-world data drastically challenges the single-domain assumptions of existing AL algorithms.

Currently, multi-domain AL lacks standard benchmark datasets, especially for large-scale, in-the-wild domain-based data \cite{he2022multidomain}.
To facilitate ongoing research,
we introduce CLIP-GeoYFCC. 
Built atop the GeoYFCC dataset \cite{geoyfcc}, the new dataset contains 1,146,768 images from 18,814 classes across 6 domain. 
It is designed to be comparable to Domainnet \cite{domainnet}, one of the largest domain-based image datasets, in terms of size, number of domains, and number of classes.
Unlike other existing domain-based datasets like Domainnet \cite{domainnet, Office-Caltech, Office-Home, MNIST}, CLIP-GeoYFCC classifies domains based on geographical origins of images, instead of on image genres.
We use both CLIP-GeoYFCC and Domainnet for our benchmark as they represent different types of domains and form a more robust test bed for domain-based algorithms (\cref{fig:sample_pictures}).

Leveraging the benchmark, we compare multiple domain allocation strategies, integrated with standard AL methods,
on three different metrics: ambient (in-distribution), mean-group, and worst-group accuracies (\cref{sec:metrics}). 
While there exist works that consider these metrics separately \cite{hierarchical, Abernethy2022minmax, Shekhar2021} and optimize algorithms for them individually, 
our benchmarking brings two unique insights. 
First, through evaluation on different types of domain-based datasets, we identify that algorithms performing best on one dataset may perform sub-optimally on others (\cref{sec:optimal}). 
This calls for testing algorithms on diverse domain-based datasets to ensure robustness. 
Second, we find that algorithms that excel on one metric often suffer on other metrics (\cref{sec:tradeoff}). 
Therefore, it is beneficial to evaluate the trade-off of domain allocation strategies between different metrics. 
Our benchmark thus underscores the open challenges in multi-domain AL and paves the way for future research.

%% file: sec/X_suppl.tex
\clearpage
\setcounter{page}{1}
\maketitlesupplementary

\begin{figure*}[ht]
\centering
    \begin{subfigure}{\textwidth}
        \centering
        \mysameheightbox{0.9\linewidth}{\hspace{0.2cm}%
            \foreach \i in {1,2,3,4} {%
            \includegraphics[height=\tempht]{compare_dataset/our_\i.jpeg}%
            \hspace{0.2cm}%
            }
        }
        \caption{Images labeled as ``resort'' in CLIP-GeoYFCC.}
    \end{subfigure}
    
    \begin{subfigure}{\textwidth}
        \centering
        \mysameheightbox{0.9\linewidth}{\hspace{0.2cm}%
            \foreach \i in {1,2,3,4} {%
            \includegraphics[height=\tempht]{compare_dataset/their_\i.jpeg}%
            \hspace{0.2cm}%
            }
        }
        \caption{Images labeled as one class in GeoYFCC.}
        \caption*{Corresponding labels in CLIP-GeoYFCC are ``resort'', ``scuba diving'', ``social dancers'', and ``coral reef''.}
    \end{subfigure}

    \caption{\textit{Comparison of labeling accuracy in CLIP-GeoYFCC and GeoYFCC}. The first images in each dataset are the same. Since GeoYFCC acquires image labels through hard filtering image tags, images in the same class often demonstrate different subjects. This example reflects a general problem existing in GeoYFCC.}
    \label{fig:compare_dataset}
\end{figure*}

\appendix 

\section{Definition of data efficiency}
\label{de}
To better describe the relative performance of AL algorithms compared with random sampling, we introduce data efficiency as a metric. \textbf{Data efficiency} (DE) quantifies the ratio of number of labeled samples needed by random sampling to that needed by an AL method to reach the same accuracy. DE could be calculated with respect to ambient, mean-group, and worst-group accuracies. An AL method \textbf{fails} if it has DE less than 1 because it requires more samples than random sampling to reach the same validation accuracy. 

Suppose at query round $r$, an active learning method $\Gamma$ and random sampling $R$ achieve validation accuracies $\Gamma(r)$ and $R(r)$ respectively. Data efficiency of $\Gamma$ at query round $r$ is defined as
\[
    DE_r(\Gamma) = \frac{\argmin_{\hat{r}} R(\hat{r}) \geq \Gamma(r)}{r}.
\]
 
\section{Two-step multi-domain active learning}
\label{sec:alg}
See \cref{alg:cap} for two-step AL algorithm where domain allocation strategy first determies per-domain sampling budgets and then instance-level query strategy is applied separately to each domain. 

\begin{algorithm}
\caption{Two-step Multi-domain Active Learning}
\label{alg:cap}
\begin{algorithmic}
\Require Neural network model $M$, unlabeled training pool $U = \{x_i\}$, a small labeled seed set $S = \{(x_i, y_i)\}$. Per-domain validation sets $V_j$'s. Per-round budget $m$, total rounds $R$
\State $L \gets S$ \Comment{Labeled training set}
\State $M \gets \text{train}(L)$ \Comment{train model $M$}
\For{$r = 1$ to $R$} 
\For{$j=1$ to $N$}
    \State $\text{metrics}_j \gets \text{evaluate}(M, V_j)$
\EndFor
\State \(\triangleright\) Apply domain allocation strategy $D$,
\State \(\triangleright\) $m_j^r$ is number of queries for domain $j$
\State $(m_1^r, ..., m_N^r)\gets D(\text{metrics}, m)$ 
\For{$j = 1$ to $N$} 
    \State \(\triangleright\) Apply instance-level query strategy $Q$
    \State $U_j^r \gets Q(U_j \setminus L_j, m_j^r)$
    \State $L_j^r\gets\text{label}(U_j^r)$
	\State $L_j \gets L_j \cup L_j^r$
\EndFor	
\State $L^r \gets \bigcup_{j=1}^{N} L_j^r$
\State $M \gets \text{train}(L^r)$
\EndFor

\end{algorithmic}
\end{algorithm}

\section{Additional filtering steps for CLIP-GeoYFCC}
\label{app:filter_geoyfcc}

\begin{figure}
    \centering
    \includegraphics[width = 0.85\columnwidth]{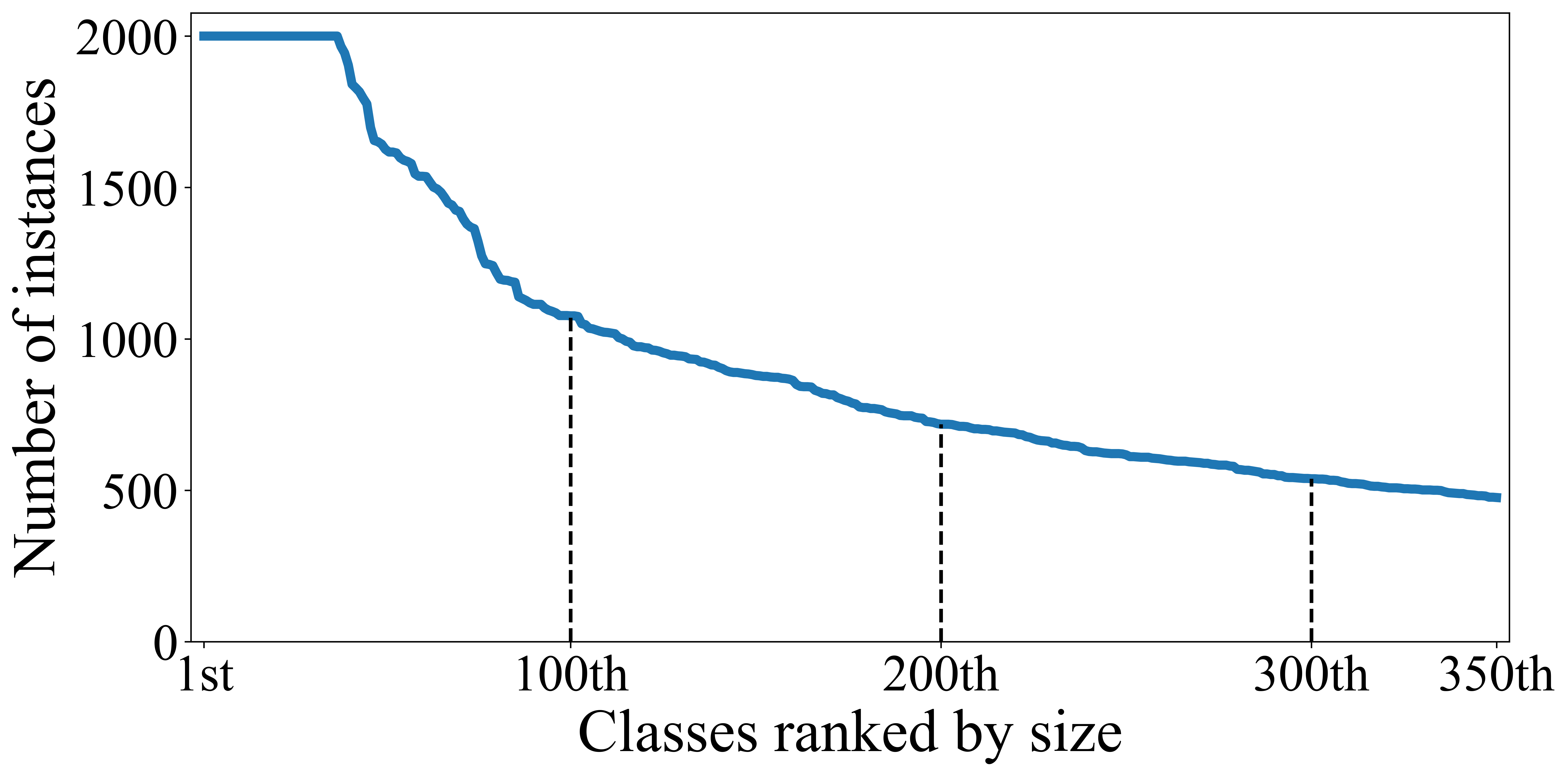}
    \caption{\textit{Sizes of classes in CLIP-GeoYFCC-350}. 
    }
    \label{fig:class_count}
\end{figure}

To match the design of Domainnet and to further clean up the dataset, we rank samples in CLIP-GeoYFCC by their similarity scores with labels matched by CLIP and select the top 1M samples. 
The domain of Oceania is removed because it accounts for less than 4\% of the total images. 
To ensure that each class has sufficient samples for training, we select the 350 largest classes where each class has at least 450 samples (\cref{fig:class_count}). 
For any class with more than 2k samples, we randomly sub-sample to limit it to 2k samples. 

Our CLIP-GeoYFCC has significantly improved the accuracy of labeling compared to the original GeoYFCC \cite{geoyfcc}. In \cref{fig:compare_dataset}, we show an example comparing the accuracy of class categorization in CLIP-GeoYFCC and GeoYFCC. The CLIP-GeoYFCC dataset and the source code will be released on github. 

\section{Uncertainty-based sampling}
\label{sec:margin_best}

\begin{table*}
\newcolumntype{Y}{c}
\centering\setlength{\tabcolsep}{2pt}
\begin{tabularx}{\textwidth}{
l|YYY|YYY|YYY|YYY}
\toprule
\textbf{Method} &
  \multicolumn{3}{c|}{\text{GeoYFCC-45}} &
  \multicolumn{3}{c|}{\text{GeoYFCC-350}} &
  \multicolumn{3}{c|}{\text{Domainnet-40}} &
  \multicolumn{3}{c}{\text{Domainnet-345}}\\
\midrule
uniform entropy            & 62.45 & 64.72 & \textbf{58.76}         & 42.04 & 44.94 & 37.54   & 68.88 & 65.24 & 38.26  & 45.34 & 41.85 & 21.14   \\
uniform least-confidence   & 61.51 & 63.79 & 57.52                  & 42.11 & 44.99 & 37.94   & 69.43 & 65.02 & 34.20  & 45.63 & 42.30 & 20.80   \\
uniform margin             & \textbf{62.56} & \textbf{64.99} & 58.67   & \textbf{43.08} & \textbf{46.23} & \textbf{38.68}   & \textbf{70.20} & \textbf{65.92} & \textbf{40.87}  & \textbf{46.93} & \textbf{43.62} & \textbf{22.98} \\
\bottomrule
\end{tabularx}
\smallskip
\caption{\textit{Comparison of accuracies of instance-level query methods under uniform domain allocation (\%)}. For each dataset, the three columns correspond to ambient accuracy, mean-group accuracy, and worst-group accuracy evaluated at the 40th round of query. Best results are shown in \textbf{bold}.}
\label{tab:margin_best}
\end{table*}

In each query round, uncertainty sampling selects the $m$ samples in $U$ with the highest $q(x; M)$ scores, where $q(x; M)$ denotes the uncertainty measure for instance $x$ under model $M$.

\paragraph{Margin sampling.} Margin sampling \cite{margin1} chooses samples with the smallest difference between the first and second most probable classes:
\[
q_{MA}(x;M) = - [p (y^* | x; M) - \max_{y_k \neq y^*} p(y_k | x; M)],
\]
where $y^* = \argmax_y p(y | x; M)$. 

\paragraph{Least confidence sampling.} Least confidence sampling \cite{LC} queries instances whose best labeling is the least confident:
\[
q_{LC}(x;M) = - \max_k p(y_k | x; M).
\]

\paragraph{Entropy sampling.} Entropy sampling \cite{entropy} chooses examples that maximize the entropy of the model’s predictive distribution:
\[
q_{ENT}(x;M) = - \sum_k p(y_k | x; M) \log p(y_k | x; M).
\]
 
Among the three uncertainty-based methods, margin sampling has been shown to be the best-performing method and it even outperforms other clustering-based methods \cite{bahri2022margin, gissin2019discriminative, ALbert}. In \Cref{tab:margin_best}, we compare the three types of uncertainty-based sampling methods as instance-level query strategies on CLIP-GeoYFCC and Domainnet. To ablate for the effect of different domain-level sampling frequencies, all three methods use the uniform domain allocation strategies. When per-domain labeling budgets are controlled, margin sampling achieves the best performances across datasets and evaluation metrics except on CLIP-GeoYFCC-45, where it fails by a small margin on worst-group accuracy. 

\section{Domain allocation strategies}
\label{sec:domain_allocation}
We benchmark four domain allocation strategies that are listed in the order of their aggressiveness towards selecting samples from the worst group. 

\paragraph{Uniform allocation.} This strategy evenly distributes the labeling budget between all domains:
\[
m_i^r = \frac{m}{N}.
\]

\paragraph{Error-proportional allocation.}
This strategy allocates the labeling budget proportional to the error rates in different domains:
\[
m_i^r = \frac{\varepsilon_i^r}{\sum_j \varepsilon_j^r} m,
\]
where $\varepsilon_i^r$ denotes the error rate of the model in round $r$ on the validation set for domain $i$.

\paragraph{Loss-exponential allocation.}
In this strategy, the labeling budget is allocated proportional to the exponent of the loss:
\[
m_i^r = \frac{e^{l_i^r}}{\sum_j e^{l_j^r}} m,
\]
where
$l_i^r$ denotes the average cross entropy \emph{training loss} at round $r$ on domain $i$.

\paragraph{Worst-group allocation.} This strategy assigns all labeling budget to the domain with the lowest validation accuracy:
\begin{equation*}
    \begin{cases}
        m & \displaystyle i = \argmax_j \varepsilon_j^r \\
        0 & \displaystyle i \neq \argmax_j \varepsilon_j^r.
    \end{cases}
\end{equation*}

\begin{figure*}
    \centering
    \hfill
    {
        \begin{subfigure}{0.45\textwidth}
            \includegraphics[width=\linewidth]{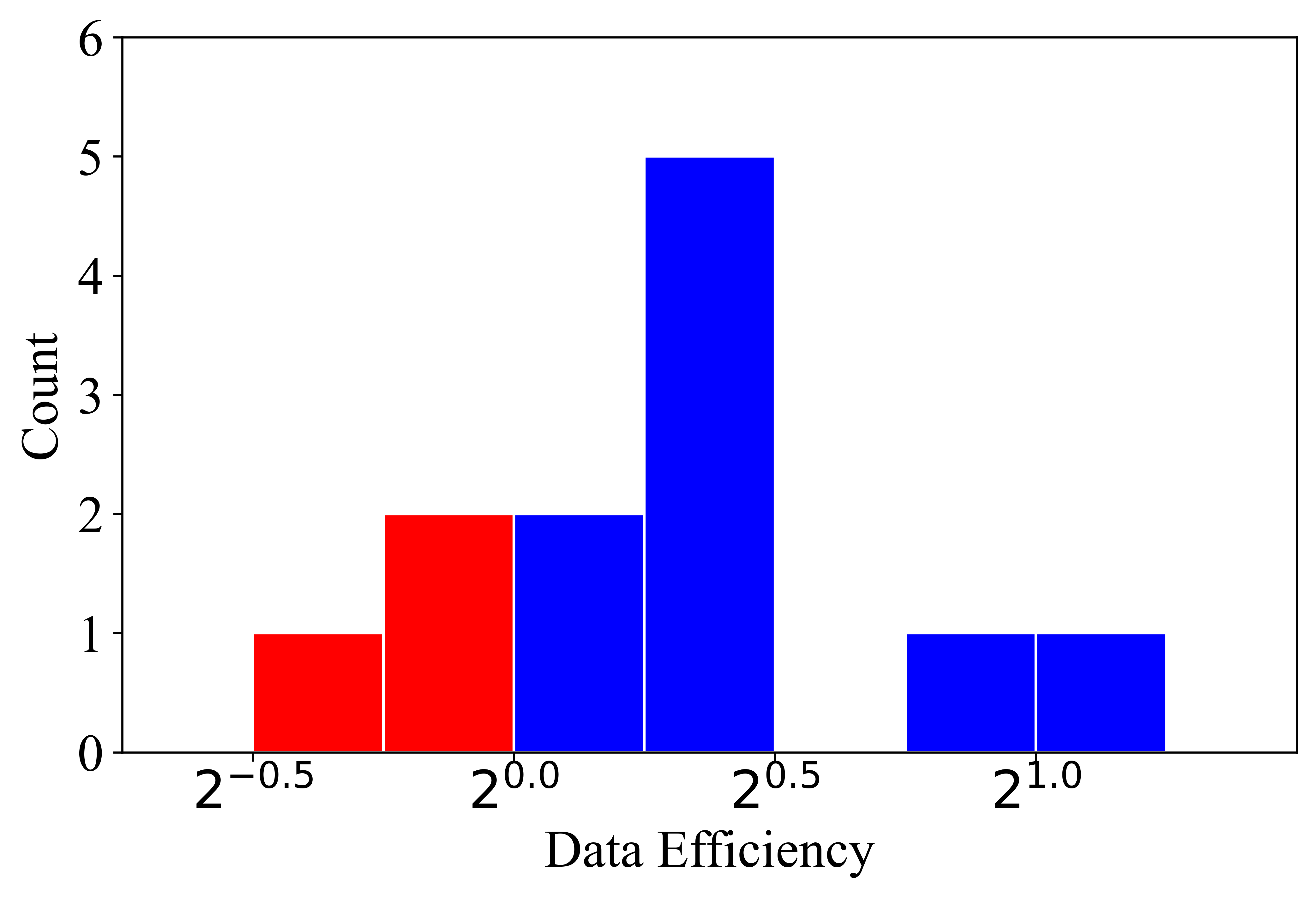}
            \caption{margin sampling}
        \end{subfigure}\hfill
    }
    {
        \begin{subfigure}{0.45\textwidth}
            \includegraphics[width=\linewidth]{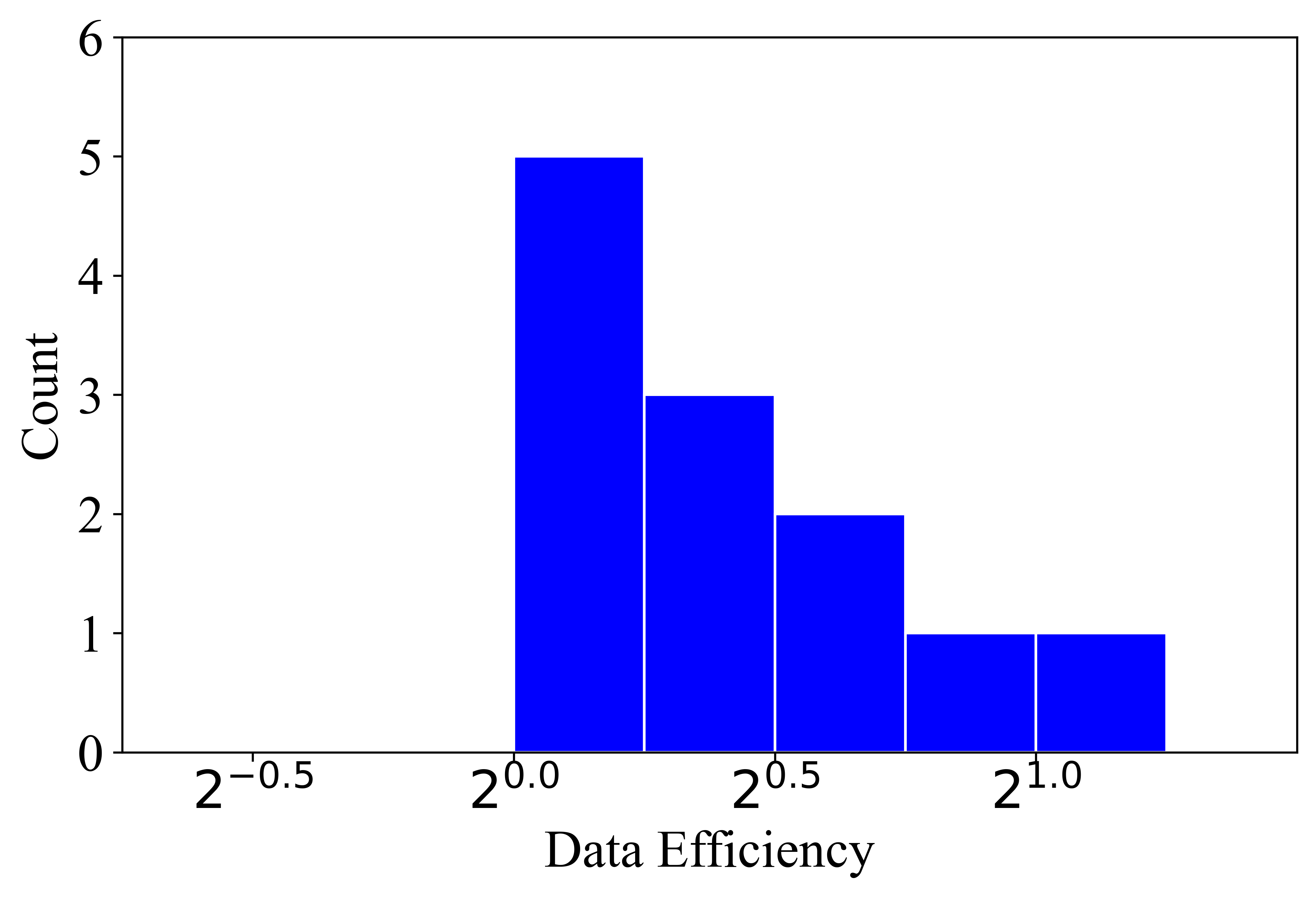}
            \caption{threshold margin}
        \end{subfigure}\hfill
    } 
    \caption{\textit{Comparing data efficiencies of margin sampling and threshold-margin sampling on Domainnet-40 domain compositions}. Data efficiencies are calculated with respect to ambient accuracy after 20 rounds of query.
    The $x$ axis is drawn in log scale.}
    \label{fig:thresholdmargin}
\end{figure*}

\section{Experiment details}
\label{sec:experiment_details}

For all experiments, we use ResNet-18 neural network pre-trained on ILSVRC12 \cite{imagenet} as the model. A new model is trained from scratch after every query in order to sample the next batch of unlabeled data. 

For experiments on CLIP-GeoYFCC-45 and Domainnet-40, the active learning process starts with a labeled seed set of 1000 samples and the labeling budget per query round is 50. For experiments on CLIP-GeoYFCC-350 and Domainnet-345, the active learning process starts with a labeled seed set of 10000 samples and the labeling budget per query round is 500.

\section{Threshold-margin}
\label{sec:threshold}

\subsection{Threshold-margin sampling}

We propose threshold-margin,
a \emph{simple}, \emph{scalable}, and \emph{hyperparameter-free} active learning strategy 
that adaptively selects samples with high uncertainties.
Threshold-margin randomly selects points with margin scores (\cref{sec:margin_best}) above a threshold,
where the threshold is automatically re-adjusted after every batch update
using a held-out validation set.
Without using any domain information, this simple scheme improves upon margin sampling over ambient accuracy
while offering comparable performances on mean-group and worst-group accuracies.

To determine the threshold after every batch update, we run model $M$ on the validation set $\{(x_i, y_i)\}_{i=1}^n$ to acquire margin scores $\{q_{MA}(x_i;M)\}_{i=1}^n$ and predicted labels $\{\hat{y}_i\}$. Threshold $t$ is determined such that 
\[
\sum_{i=1}^n \mathbbm{1}\{q_{MA}(x_i;M) > t\} = \sum_{i=1}^n \mathbbm{1}\{\hat{y}_i \neq y_i\}.
\]
The idea is that the expected number of points receiving margin scores higher than $t$ matches the error of the model on the validation data. 

This thresholding method is introduced in \cite{ATC} to predict target domain accuracy using only labeled source data and unlabeled target data. \cite{ATC} calculates a threshold on labeled source data and predict model's accuracy on unlabeled target data as the proportion of unlabeled samples with scores above the threshold. This method has demonstrated state-of-the-art performance in out-of-domain accuracy prediction. 
In active learning, we leverage the threshold to determine data points that are likely to be errors. 

After calculating the threshold $t$, threshold-margin randomly selects unlabeled datapoints with scores above the threshold with equal probability. We run model $M$ on unlabeled data $\{x_i\}_{i=1}^K$ and acquire margin scores $\{q_{MA}(x_i;M)\}_{i=1}^K$. Probability of selecting sample $x_i$ in a single query is  
\[
ps (x_i; M, t) \propto \mathbbm{1}\{q_{MA}(x_i;M) > t\}.
\]
The selection procedure could be modified to be more computation efficient.
The algorithm can randomly sample datapoints and select all samples with scores above the threshold.
The process terminates as soon as enough datapoints have been selected.
This can be far more efficient than running inference on the entire unlabeled pool.

Compared with margin sampling that selects instances with lowest margin scores, threshold-margin sampling balances hard examples with diversity and achieves some improvement on ambient accuracies. 
This method could be applied to single-domain active learning or be used as instance-level query strategy as it does not require domain information.

\subsection{Threshold-group-margin}

Threshold-group-margin is a modification of threshold-margin that incorporates domain information and balances mean-group accuracy with worst-group accuracy. 
To account for domain imbalance, threshold-group-margin scales the probability of selecting each unlabeled sample inversely with its corresponding domain size.
More concretely, suppose unlabeled samples $x_i$ comes from domain $j$ and $N_j$ denotes the number of unlabeled samples from domain $j$. Then the probability of selecting sample $x_i$ in a single query is
\[
ps (x_i; M, t) \propto \mathbbm{1}\{q_{MA}(x_i;M) > t\} \frac{1}{N_j}.
\]

\subsection{Performance of threshold-margin}

On average across different datasets, threshold-margin provides small gain on ambient accuracy compared to margin sampling (\cref{tab:thmargin_table}). However, we note that threshold-margin does not perform better than margin sampling on every dataset. In \cref{fig:thresholdmargin}, we compare threshold-margin and margin in terms of their data efficiencies with respect to ambient accuracy. We use the same 12 domain compositions of Domainnet-40 as those in \cref{fig:domain_effect}. The major benefit of threshold-margin is that its data efficiency is never below 1 (worse than random sampling) and is more robustness to different domain compositions. 

\begin{table}[ht]
\centering
\begin{tabularx}{0.47\textwidth}{l YYY}
\toprule
\textbf{method} &
  \textbf{amb} &
  \textbf{mean} &
  \textbf{worst} \\
  \midrule
margin              & 58.66         & 57.41          & 48.92         \\
threshold-margin   & \textbf{59.06} & 57.74          & 49.29         \\
threshold-group-margin &
  \underline{58.75} &
  \textbf{58.57} &
  49.58 \\
uniform margin     & 58.49          & \underline{58.37} & 48.70          \\
error-prop margin & 58.24          & 58.17          & 49.37          \\
loss-exp margin   & 57.85          & 57.85         & \underline{50.53}          \\
worst-group margin      & 55.97          & 55.32          & \textbf{52.52} \\
\bottomrule
\end{tabularx}
\caption{\textit{Aggregate accuracy (\%) of different domain allocation methods}. Reported accuracies are averaged over the experiments on CLIP-GeoYFCC-45, CLIP-GeoYFCC-350, Domainnet-40(4), and Domainnet-345.
The \textbf{highest} and \underline{second highest} accuracies are shown in \textbf{bold} and \underline{underline}, respectively.
Readers are referred to \cref{fig:four_datasets_thmargin} for results on individual datasets.}
\label{tab:thmargin_table}
\end{table}

\begin{figure*}[ht]
    \centering
    \includegraphics[width = 0.95\textwidth]{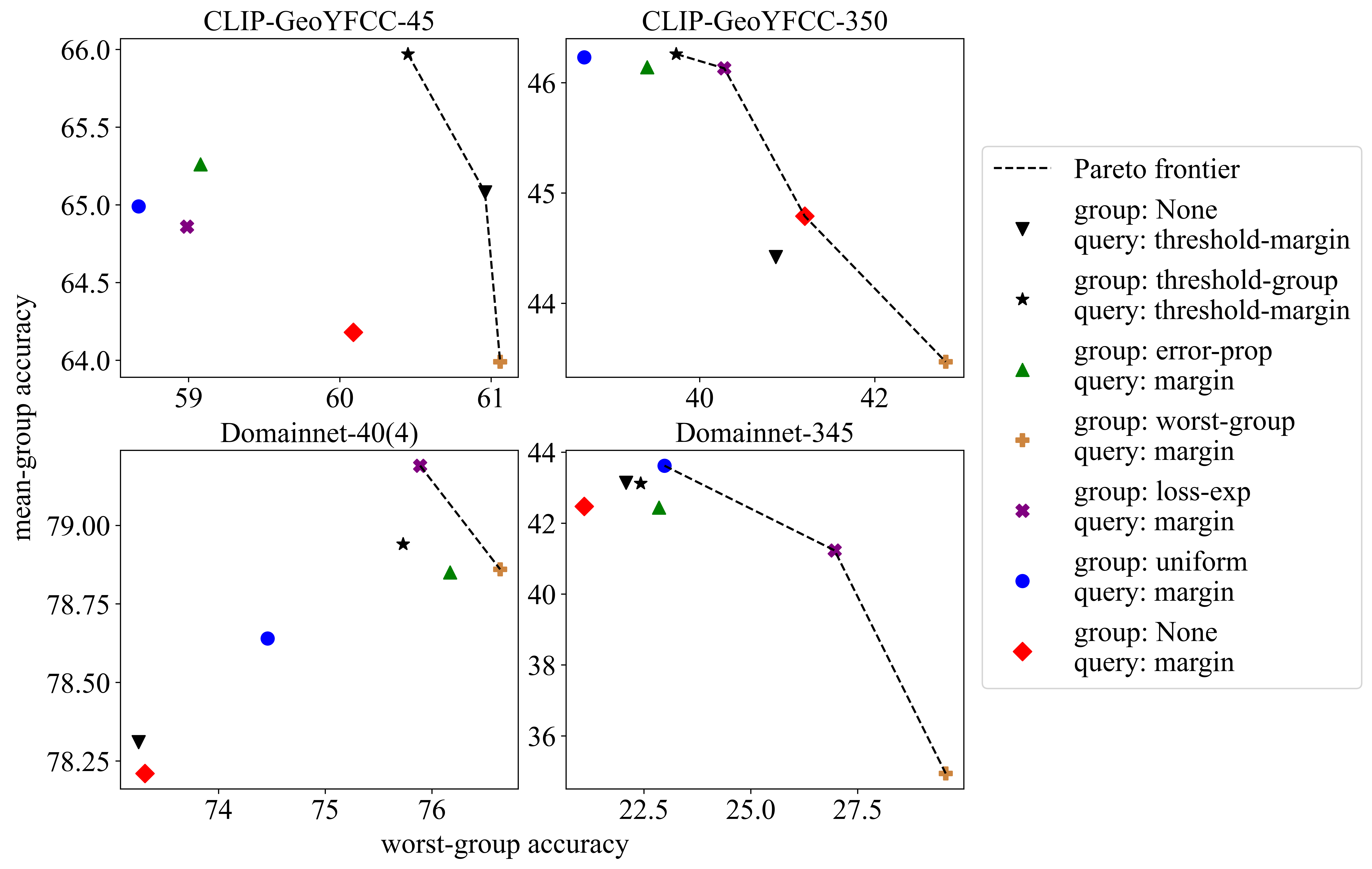}
    \caption{\textit{Comparing different domain allocation methods on CLIP-GeoYFCC-45, CLIP-GeoYFCC-350, Domainnet-40(4), and Domainnet-345}. Domainnet-40(4) consists of 4 domains (clipart, painting, real, sketch) out of the 6 domains in Domainnet-40. 
    Accuracies are evaluated after 40 rounds of query. 
    The dashed line represents the Pareto frontier \cite{Martinez2020pareto} between the mean-group and worst-group accuracies.
    }
    \label{fig:four_datasets_thmargin}
\end{figure*}

\subsection{Limitations of threshold-margin}

While threshold-margin exhibits some benefits on ambient accuracy, we acknowledge several limitations of our current investigation of this method. Firstly, when calculating threshold after every batch update, our current algorithm uses a fixed held-out validation set that is randomly sampled from the dataset and labeled at the beginning. We have not investigated updating validation set along the AL process and have not evaluated the impact of initial validation set on the performance of threshold-margin. Secondly, we are not certain about the exact reason why threshold-margin performs well and demonstrates more robustness in practice. One hypothesis is that margin sampling tends to select instances with similar types of errors and threshold-margin increases selection diversity but more rigorous ablation study is needed. Lastly, performance of threshold-margin is not consistent across datasets. For example, threshold-margin significantly outperforms margin on CLIP-GeoYFCC-45 but only brings marginal improvements on Domainnet (\cref{fig:four_datasets_thmargin}).